\documentclass{article}
\usepackage{ijcai20}

\usepackage{times}
\usepackage{soul}
\usepackage{url}
\usepackage[hidelinks]{hyperref}
\usepackage[utf8]{inputenc}
\usepackage[small]{caption}
\usepackage{graphicx}
\usepackage{amsmath}
\usepackage{amsthm}
\usepackage{booktabs}
\usepackage{algorithm}
\usepackage{algorithmic}
\usepackage{subfigure}
\usepackage{natbib}
\usepackage{amsfonts}
\title{Balanced One-shot Neural Architecture Optimization}

\author{
Renqian Luo$^1$\and
Tao Qin$^2$\and
Enhong Chen$^1$
\affiliations
$^1$University of Science and Technology of China\\
$^2$Microsoft Research Asia\\
\emails
lrq@mail.ustc.edu.cn,
taoqin@microsoft.com,
cheneh@ustc.edu.cn
}

\begin{document}

\maketitle

\begin{abstract}
The ability to rank candidate architectures is the key to the performance of neural architecture search~(NAS). One-shot NAS is proposed to reduce the expense but shows inferior performance against conventional NAS and is not adequately stable. We investigate into this and find that the ranking correlation between architectures under one-shot training and the ones under stand-alone full training is poor, which misleads the algorithm to discover better architectures. Further, we show that the training of architectures of different sizes under current one-shot method is imbalanced, which causes the evaluated performances of the architectures to be less predictable of their ground-truth performances and affects the ranking correlation heavily. Consequently, we propose \emph{Balanced NAO} where we introduce balanced training of the supernet during the search procedure to encourage more updates for large architectures than small architectures by sampling architectures in proportion to their model sizes. Comprehensive experiments verify that our proposed method is effective and robust which leads to a more stable search. The final discovered architecture shows significant improvements against baselines with a test error rate of 2.60\% on CIFAR-10 and top-1 accuracy of 74.4\% on ImageNet under the mobile setting. Code and model checkpoints will be publicly available. Code is available at \url{github.com/renqianluo/NAO\_pytorch}.
\end{abstract}

\section{Introduction}
Neural architecture search~(NAS) aims to automatically design neural network architectures. Recent NAS works show impressive results and have been applied in many tasks, including image classification~\cite{nas,nasnet}, object detection~\cite{nasfpn}, super resolution~\cite{chu2019fast}, language modeling~\cite{enas}, neural machine translation~\cite{evovledtransformer} and model compression~\cite{autoslim}. Without loss of generality, the search process can be viewed as iterations of two steps: (1) In each iteration, the NAS algorithm generates some candidate architectures to estimate and the architectures are trained and evaluated on the task. (2) Then the algorithm learns from the evaluation results of the candidate architectures. In principle, it needs to estimate each generated candidate architecture by training it until convergence, therefore the estimation procedure is resource consuming.

\begin{figure}[ht]
\centering
\includegraphics[width=1.0\columnwidth]{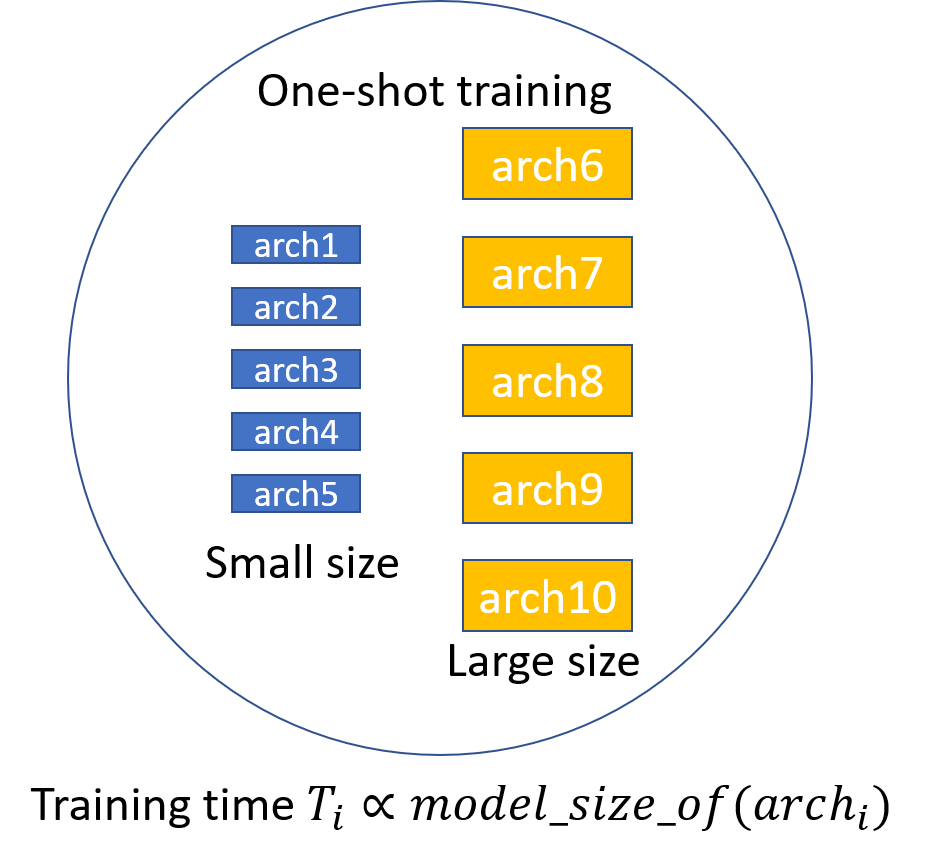}
\caption{Balanced One-shot Neural Architecture Optimization. Each architecture is sampled in proportion to its model size during training in the search procedure, to balance architectures of different sizes.}
\label{fig:bnao}
\end{figure}

One-shot NAS was proposed~\cite{oneshot,enas} to reduce the resources, by utilizing a supernet to include all candidate architectures in the search space and perform weight inheritance between different architectures. However, the architectures discovered by such one-shot NAS works show inferior performance to conventional NAS which trains and evaluates each architecture individually. Meanwhile, the stability is weak that the architectures generated by the algorithms do not always achieve the performance reported originally~\cite{random}. We conduct experiments on ENAS~\cite{enas}, DARTS~\cite{darts} and NAO-WS~\cite{nao}, and find that the stability is weak and the performance is inferior to conventional NAS~\cite{nasnet,amoebanet,nao}.

\begin{figure}[ht]
	\centering
	\subfigure[Conventional NAS]{
		\label{fig:conventional nas}
		\includegraphics[width=1.0\columnwidth]{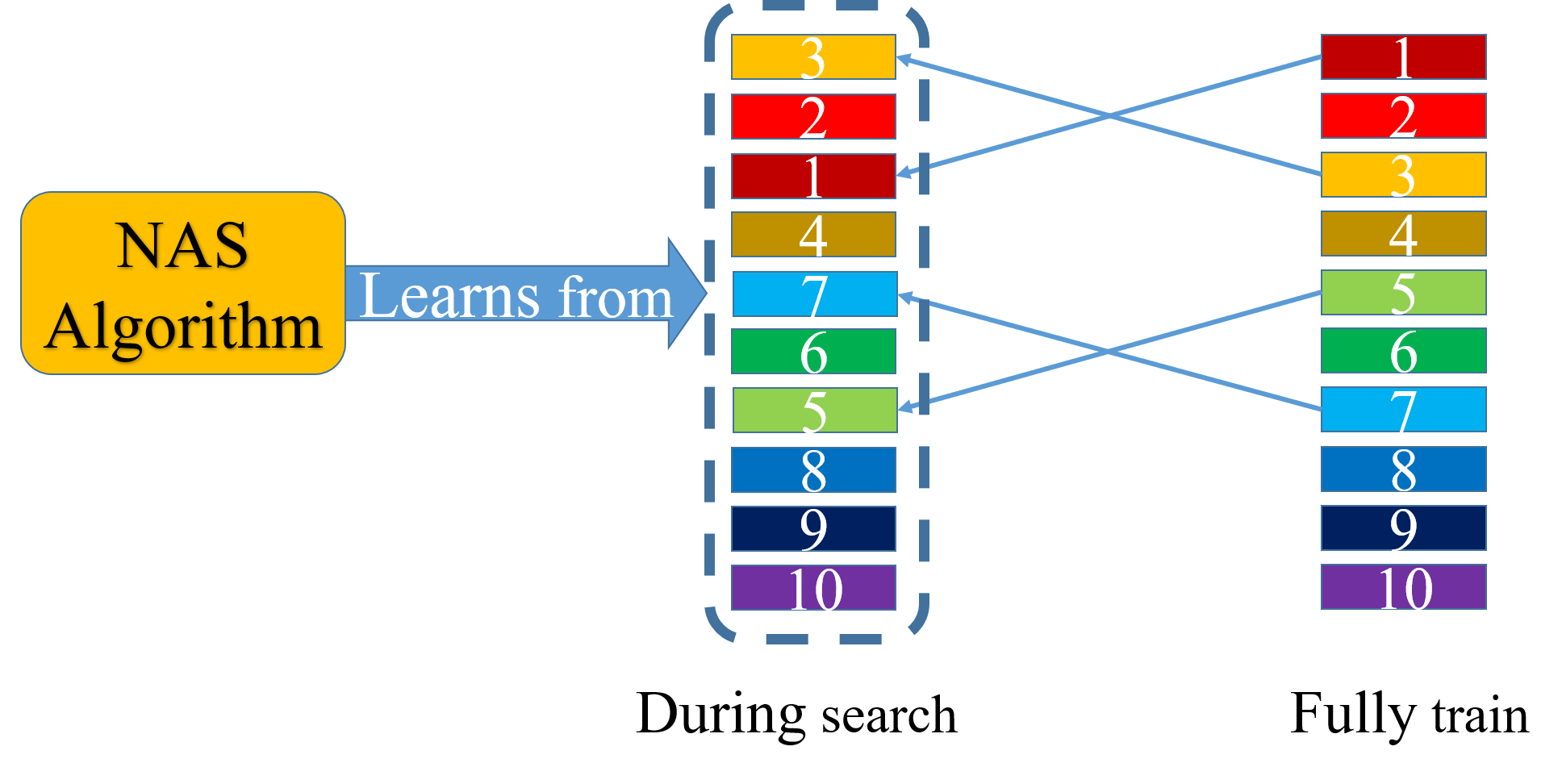}
	}
	\subfigure[One-shot NAS]{
		\label{fig:one-shot nas}
		\includegraphics[width=1.0\columnwidth]{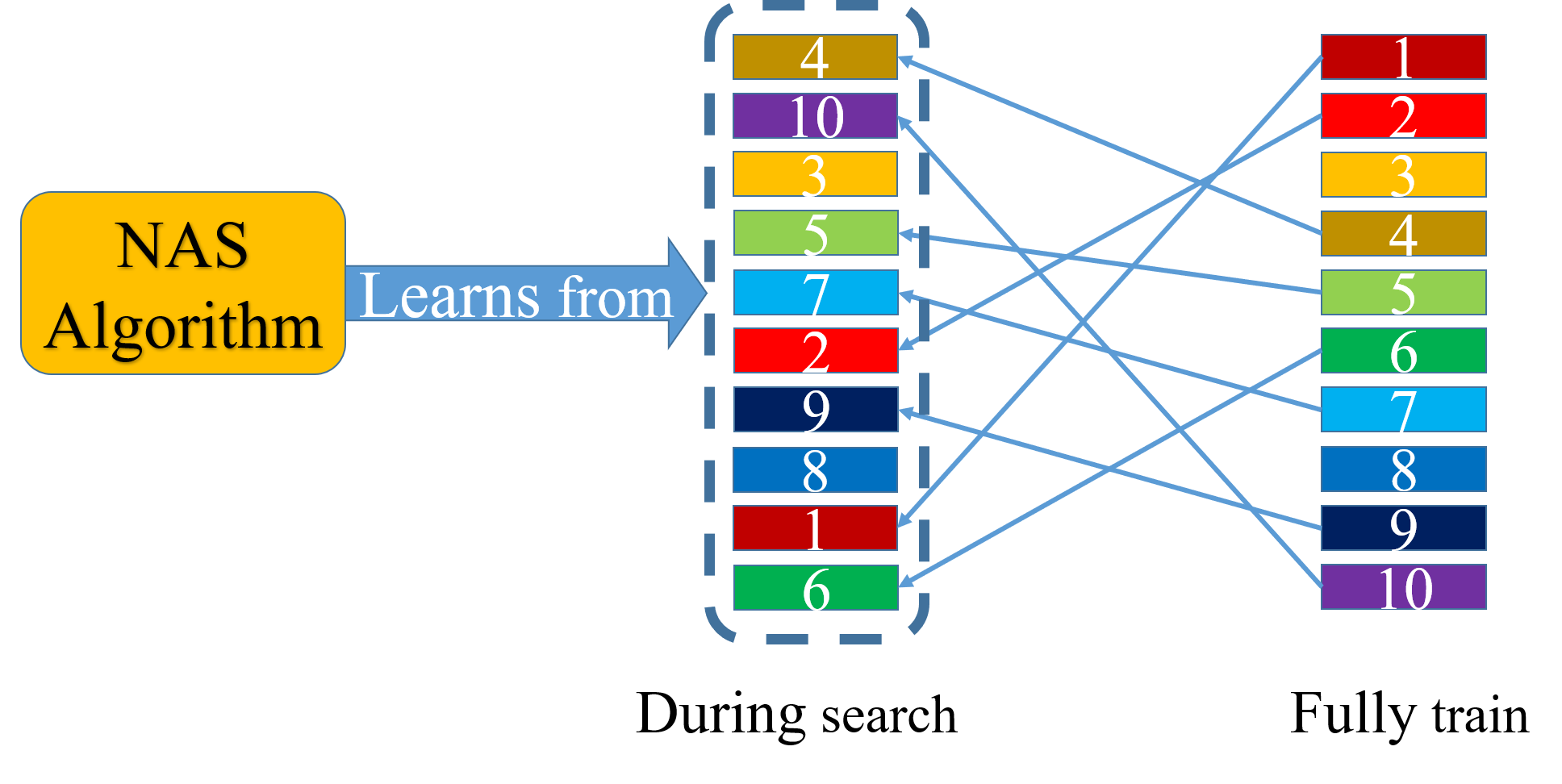}
	}
\caption{Illustration of ranking correlation between architectures in neural architecture search. Rectangles with different colors and numbers represent different architectures ranked by performance. (a) In conventional NAS, the ranking correlation between architectures during search and final fully trained ones is well preserved with slight bias. (b) In one-shot NAS, the ranking correlation is hardly preserved.}
\label{fig:nas framework}
\end{figure}

During the search, the algorithm aims to differentiate the candidate architectures and figures out relatively good architectures among the all. Then the key to the algorithm is the ability to rank these candidate architectures. Then we should expect that the ranking of the architectures during final full training to be preserved when they are trained under such one-shot method during the search, so the algorithm can learn well. We investigate into the one-shot training with the weight-sharing mechanism and find that the ranking correlation of the architectures is not well preserved under such one-shot training, which misleads the algorithm to effectively and robustly find an optimal architecture, as shown in Fig.~\ref{fig:nas framework}. Architectures discovered could perform relatively well when trained in such one-shot method, without guarantee to perform well when fully trained stand-alone.

Under one-shot training, thousands of candidate architectures are included in a supernet and one architecture~(or one dominant architecture) is sampled at each step during the training to update its parameters. The average training time of each architecture is handful and the optimization is insufficient, where the validation accuracy of the architecture is less representative of its ground-truth performance in final training. We show that the ranking correlation of the architectures is weak under one-shot training, and the ranking correlation increases as the training time increases. 

Since small architectures are easier to train than big architectures, given handful training steps and insufficient optimization, small architectures tend to perform well than big architectures, which is not consistent in the final training. Therefore the training of the candidate architectures is imbalanced, and the ranking correlation is weak, which is demonstrated by our experiments. This misleads the search algorithm to effectively find good architecture and the algorithms are observed to produce some small architectures that generalize worse during the final training.

Based on such insight, we propose balanced NAO where we introduce balanced training to encourage large architectures for more updates than small architectures during the one-shot training of the supernet, depicted in Fig.~\ref{fig:bnao}. We expect the training time of an architecture to be proportional to its model size. Specifically, we sample a candidate architecture in proportion to its model size, instead of uniformly random sampling. With this proportional sampling approach, the expected training time of the architecture associates with its size, and experiments show that the ranking correlation increases.

Extensive experiments on CIFAR-10 demonstrate that:~(1) Architectures discovered by our proposed method show significant improvements compared to baseline one-shot NAS methods, with a test error rate of 2.60\% on CIFAR-10 and top-1 accuracy of 74.4\% on ImageNet.~(2) Our proposed method generates architectures with more stable performance and lower variance, which brings better stability.

\section{Related Work}
Recent NAS algorithms mainly lie in three lines: reinforcement learning~(RL), evolutionary algorithm~(EA) and gradient based.~\cite{nas} firstly proposed to search neural network architectures with reinforcement learning~(RL) and achieved better performances than human-designed architectures on CIFAR-10 and PTB. ~\cite{amoebanet} proposed AmoebaNet using regularized evolutionary algorithm.~\cite{nao} maps the discrete search space to continuous space and search by gradient descent. These works achieve promising results on several tasks, however they are very costly.

To reduce the large amount of resources conventional NAS methods require, one-shot NAS was proposed.~\cite{oneshot} proposed to include all candidate operations in the search space within a supernet and share parameters among different candidate architectures. ENAS~\cite{enas} leverages the idea of weight sharing and searches by RL. DARTS~\cite{darts} searched based on the supernet through a bi-level optimization by gradient descent. NAO~\cite{nao} also incorporates the idea of weight sharing. ProxylessNAS~\cite{proxylessnas} binaries architecture associated parameters to reduce the memory consumption in DARTS and can directly search on large-scale target task. SNAS~\cite{snas} trains the parameters of the supernet and architecture distribution parameters in the same round of back-propagation. However, such supernet requires delicate design and needs to be carefully tuned.

\section{Analyzing One-shot NAS}
\subsection{Stability of Baselines}
\label{sec:analysis}
\begin{table}[ht]
\centering
\begin{tabular}{cccc}
\hline
Architectures & ENAS & DARTS & NAO-WS \\
\hline
Original & 2.89 & 2.83 & 2.93 \\
\hline
network 1  & 3.09 &3.00 & 2.90 \\
network 2  & 3.00 &2.82 & 2.96 \\
network 3  & 2.97 &3.02 & 2.98 \\
network 4  & 2.79 &2.96 & 3.02 \\
network 5  & 3.07 &3.00 & 3.05 \\
\hline
\end{tabular}
\caption{Evaluation of the stability of three recent one-shot NAS algorithms. The first block shows the test performance reported by the original authors. The second block shows the test performance of architectures we got using the algorithms. We ran the search process for $5$ times, and evaluated the $5$ network architectures generated. All the models are trained with cutout~\cite{cutout} and the results are test error rate on CIFAR-10. We follow the default settings and training procedure in the original works.}
\label{analysis:baseline}
\end{table}
Firstly, we study the stability of three previous popular one-shot NAS algorithms, ENAS~\cite{enas}, DARTS~\cite{darts} and NAO~\cite{nao}, as many one-shot NAS works are based on these three works. For NAO, we use NAO-WS~(NAO with weight sharing), which leverages the one-shot method. We use the official implementation of ENAS\footnote{https://github.com/melodyguan/enas.git}, DARTS\footnote{https://github.com/quark0/darts.git} and NAO-WS\footnote{https://github.com/renqianluo/NAO\_pytorch.git}. We searched and evaluated on CIFAR-10 dataset, and adopted default settings and hyper-parameters by the authors, if not stated explicitly in the following context. We ran the search process for $5$ times, and evaluated the discovered architectures. The results are shown in Table~\ref{analysis:baseline}.

We can see that for all the three algorithms, among the $5$ network architectures we evaluated, at least one of them could achieve the performance as reported in the original paper. However the results are unstable. ENAS achieved $2.79$\% in the fourth run which is on par with the result reported in the paper~($2.89$\%), while the other four runs got inferior results with about $0.2$ drop. DARTS achieved $2.82$\% in the second run which is on par with the result reported originally~($2.83$\%), but got around $3.0$\% in the other four runs with about $0.2$ drop. NAO-WS achieved $2.90$\% and $2.96$\% on the first and the second architectures which are on par with the result reported by the authors~($2.93$\%), but got around $3.0$\% on the other three architectures with also about $0.1$ drop. Moreover, all the results are inferior to the performance achieved by conventional NAS algorithms which do not leverage one-shot method~(e.g., AmoebaNet-B achieved $2.55$\%, NAONet achieved $2.48$\%).

In the search process, the ability of the performance of an architecture under one-shot training~(noted as \emph{one-shot performance}) to predict its performance under final full training~(noted as \emph{ground-truth performance}) is vital. Further, as the algorithm is practically supposed to learn to figure out relatively better architectures among the candidate architectures according to their one-shot performance, the \emph{ranking correlation} between candidate architectures during one-shot training and full training is the key to the algorithm. That is, if one architecture is better than another architecture in terms of ground-truth performance, we would expect that their one-shot performance also preserves the order, no matter what concrete values they are. Since the search algorithms of most one-shot NAS works still hold RL, EA or gradient based optimization that are almost the same to the conventional NAS~(e.g., NAO has both regular version and one-shot version while the latter shows inferior performance), we conjecture that the challenge is in the one-shot training mechanism.

\subsection{Insufficient Optimization}
First, we consider the issue is due to the insufficient optimization of the architectures, as is also shown by~\cite{random}. In one-shot training, the average training time of an individual architecture is surprisingly inadequate. Generally, an architecture should be trained for hundreds of epochs to converge in the full training pipeline to evaluate its performance. In conventional NAS, early-stop is performed to cut-off the high cost but each architecture is still trained for several tens of epochs, so the performance ranking of the models is relatively stable and can be predictable of their final ground-truth performance ranking. However in one-shot NAS, the supernet is usually trained in total for the time that training a single architecture needs. Since all the candidates are included, in average each architecture is trained for handful steps. Though with the help of weight-sharing and the parameters of an architecture could be updated by other architectures that share the same part, the operations could play different roles in different architectures and may be affected by other ones. Therefore, architectures are still far from convergence and the optimization is insufficient. Their evaluated one-shot performance are largely biased from their ground-truth performance. 

\begin{table}[ht]
\centering
\begin{tabular}{ccc}
\hline
Training Epochs & Pairwise accuracy(\%) \\
\hline
5  & 62 \\
10 & 65 \\
20 & 71 \\
30 & 75 \\
50 & 76 \\
\hline
\end{tabular}
\caption{Pairwise accuracy of one-shot performance and ground-truth performance under different training epochs.}
\label{analysis:time}
\end{table}
To verify this, we conduct experiments to see how training time affects the ranking correlation. We randomly generate $50$ different architectures and train each architecture stand-alone and completely, exactly follow the default settings in~\cite{nao}, and collect their performance~(valid accuracy) on CIFAR-10. Then we train the architectures using one-shot training with weight-sharing. Concretely at each step, one architecture from the $50$ architectures is uniformly randomly sampled and trained at each step. We train the architectures under one-shot training for different time~(i.e., $5$, $10$, $20$, $30$, $50$ epochs respectively), and collect their performance on CIFAR-10. 

In order to evaluate the ranking correlation between the architectures under one-shot training and stand-alone trained ones, we calculate the pairwise accuracy between the one-shot performance and the stand-alone performance of the given architectures. The pairwise accuracy metric is the same as in~\cite{nao} which ranges from $0$ to $1$, meaning the ranking is completely reversed or perfectly preserved. The results are listed in Table~\ref{analysis:time}. 

We can see that, when the training time is short~(as in the case in practical one-shot NAS), the pairwise accuracy is $62\%$ which is slightly better than random guess, meaning that the ranking is not well-preserved. The ranking of the architectures under one-shot training is weakly correlated to that of ground-truth counterpart. As the training time increases, the average number of updates of individual architecture increases and pairwise accuracy increases evidently. One may think that when training for $50$ epochs, the pairwise accuracy seems to be satisfactory and commonly the supernet is trained for hundreds of epochs. However we should notice that this is only a demonstrative experiment where we only use $50$ architectures to train the supernet. In practice, the search algorithms always need to sample thousands of architectures~(i.e., more than $10$ times) to train~\cite{enas,nao,amoebanet} so the average training time of each individual architecture is handful as limited as the setting of $5$ epochs here. This indicates that the short average training time and inadequate update of individual architecture lead to insufficient optimization of individual architecture in one-shot training and brings weak ranking correlation.

\subsection{Imbalanced Training of Architectures}
\label{sec:imbalanced}
\begin{table}[ht]
\centering
\begin{tabular}{cccc}
\hline
Model Sizes & Pairwise Accuracy(\%) \\
\hline
diverse & 62 \\
similar & 73 \\
\hline
\end{tabular}
\caption{Pairwise accuracy of one-shot performance and ground-truth performance, of architectures with similar sizes and diverse sizes.}
\label{analysis:imbalanced}
\end{table}
During multiple runs of one-shot NAS~(e.g., DARTS and NAO-WS), we find that they sometimes tend to produce small networks, which contain many parameter-free operations~(e.g., skip connection, pooling). These architectures are roughly less than $2MB$. Theoretically, this means that these small architectures should perform well on the target task during final training. However, they do not consistently perform well when fully trained and on the contrary, they generally perform worse. This is an interesting finding, meaning that during the one-shot training, these inferior small architectures tend to perform well than big architectures, so they are picked out by the algorithms as good architectures.

In general, small architectures are easier to train and need fewer updates than big architectures to converge. Given the insights above of the inadequate training and insufficient optimization, we conjecture that small networks are easier to be optimized and tend to perform better than big networks. Big networks are then relatively less-trained on the contrary compared to small networks and perform worse. Consequently, the algorithm learns from the incorrect information that small networks would perform well. In conventional NAS, where each architecture is trained for tens of epochs though with early-stopping, both small architectures and big architectures are updated relatively more adequately and their performance ranking is more stable and close to the final ground-truth ranking. 

Therefore, different architectures are imbalanced during one-shot training. To verify this and demonstrate whether this affects the ranking correlation, we conduct experiments on architectures with diverse sizes~(as in the practical case) and similar sizes. For architectures with diverse sizes, we just use the architectures and the results in the above experiment in Table~\ref{analysis:time}. Then we randomly generate other $50$ architectures with similar sizes and train them using supernet. 

From the results in Table~\ref{analysis:imbalanced}, we can see that when the architectures are of diverse sizes, the pairwise accuracy is low. When the architectures have similar model sizes, the pairwise accuracy increases. This indicates that in practical case the training of the architectures with different sizes is imbalanced and the ranking correlation is not preserved well.

\begin{table*}[h]
\centering
\begin{tabular}{ccccc}
\hline
Model                            & Test error(\%)     & Params(M)  & Methods\\
\hline
Hier-EA~\cite{hanxiao}           & 3.75               & 15.7     & EA \\
NASNet-A~\cite{nasnet}           & 2.65               & 3.3      & RL \\
AmoebaNet-B~\cite{amoebanet}     & 2.13               & 34.9     & EA \\
NAONet~\cite{nao}                & 2.48               & 10.6     & gradients \\
NAONet~\cite{nao}                & 1.93               & 128      & gradients \\
\hline
ENAS~\cite{enas}                 & 2.89               & 4.6      & one-shot + RL \\
DARTS~\cite{darts}               & 2.83               & 3.4      & one-shot + gradients \\
NAO-WS~\cite{nao}                & 2.90               & 2.5      & one-shot + gradients \\
SNAS~\cite{snas}                 & 2.85               & 2.9      & one-shot + gradients \\
P-DARTS~\cite{pdarts}            & 2.50               & 3.4      & one-shot + gradients \\
PC-DARTS~\cite{pcdarts}          & 2.57               & 3.6      & one-shot + gradients \\
BayesNAS~\cite{bayesnas}         & 2.81               & 3.4      & one-shot + gradients \\
ProxylessNAS~\cite{proxylessnas} & 2.08               & 5.7      & one-shot + gradients \\
Random~\cite{darts}              & 3.29               & 3.2      & random               \\
\hline
Balanced NAO                     & 2.60              & 3.8      & one-shot + gradients \\
\hline
\end{tabular}
\caption{Test performance on CIFAR-10 dataset of our discovered architecture and architectures by other NAS methods, including methods with or without one-shot method. The first block is conventional NAS without one-shot method. The second block is one-shot NAS methods.}
\label{tbl:cifar10}
\end{table*}

\begin{table}[h]
\centering
\begin{tabular}{cccc}
\hline
Model/Method                                    & Top-1(\%) \\
\hline
MobileNetV1~\cite{mobilenet}                    & 70.6 \\     
MobileNetV2~\cite{mobilenetv2}                  & 72.0 \\   
\hline
NASNet-A~\cite{nasnet}                          & 74.0 \\   
AmoebaNet-A~\cite{amoebanet}                    & 74.5 \\   
MnasNet~\cite{mnasnet}                          & 74.0 \\   
NAONet~\cite{nao}                               & 74.3 \\   
DARTS~\cite{darts}                              & 73.1 \\   
SNAS~\cite{snas}                                & 72.7 \\
P-DARTS(C10)~\cite{pdarts}                      & 75.6 \\
P-DARTS(C100)~\cite{pdarts}                     & 75.3 \\
\hline
Single-Path NAS~\cite{singlepathnas}            & 74.96 \\   
Single Path One-shot~\cite{singlepathoneshot}   & 74.7 \\   
ProxylessNAS~\cite{proxylessnas}                & 75.1 \\   
PC-DARTS~\cite{pcdarts}                         & 75.8 \\
\hline
Balanced NAO                                    & 74.4 \\   
\hline
\end{tabular}
\caption{Test performances on Imagenet dataset under mobile setting. The first block shows human designed networks. The second block lists NAS works that search on CIFAR-10 or CIFAR-100 and directly transfer to ImangeNet. The third block are NAS works that directly search on ImageNet}
\label{tbl:imagenet}
\end{table}

\section{Method}
We have shown that the training of the architectures in common one-shot NAS is imbalanced, then we propose an approach to balance architectures with different sizes. We mainly propose our approach based on NAO~\cite{nao} for the following two reasons: Firstly, both ENAS and NAO perform sampling from candidate architectures at each step during the search. NAO directly samples from the architecture pool while ENAS samples from the learned controller, so the sampling procedure is easier to control in NAO. Secondly, DARTS and DARTS-like works~(e.g., ProxylessNAS~\cite{proxylessnas}, SNAS~\cite{snas}) combine the parameter optimization and architecture optimization together in a forward pass which is not based on sampling architectures.

To balance the architectures, we expect to encourage more updates for large architectures over small architectures. Specifically, we expect an architecture $x_i$ to be trained for $T_i$ steps where $T_i$ is in proportional to its model size, following:
\begin{equation}
\label{eqn:timepropto}
T_i \propto model\_size\_of(x_i)
\end{equation}

One naive and simple way is to calculate the size of the smallest model noted as $x_s$ in the search space, and set a base step $T_{base}$ for it. Then for any architecture $x_i$, its number of training steps is calculated as:
\begin{equation}
\label{eqn:time}
T_i = T_{base} \times \frac{model\_size\_of(x_i)}{model\_size\_of(x_s)}
\end{equation}
However, such a naive implementation may lead to expensive cost which may run out of the search budget, and the relationship between training time and model size is not as simple as a strictly linear relationship.

Therefore, we use proportional sampling to address the problem. Specifically in NAO~\cite{nao}, during the training of the supernet, an architecture is randomly sampled from the candidate architecture population at each step. Instead of randomly sampling, we sample the architectures with a probability in proportion to their model sizes. Mathematically speaking, we sample an architecture $x_i$ from the candidate architecture pool $\mathcal{X}$ with the probability:
\begin{equation}
\label{eqn:sample}
P(x_i)=\frac{model\_size\_of(x_i)}{\sum_{x_i\in \mathcal{X}} model\_size\_of(x_i)}.
\end{equation}
If the supernet is trained for $T_t$ steps in total, then the expectation of the training steps of architecture $x_i$ is:
\begin{equation}
\label{eqn:exptime}
\mathbb{E}T_i=T_t \times \frac{model\_size\_of(x_i)}{\sum_{x_i \in \mathcal{X}} model\_size\_of(x_i)}
\end{equation}
It is proportional to the model size of $x_i$. Following this, the architectures of different sizes could be balanced during training within the supernet.

We incorporate this proportional sampling methodology into NAO~\cite{nao}, specifically the one-shot NAO version~(i.e., NAO-WS in the original paper) which utilizes weight sharing technology, and propose \emph{Balanced NAO}. The final detailed algorithm is shown in Alg.~\ref{alg:BNAO}.
\begin{algorithm}[ht]
\caption{Balanced Neural Architecture Optimization}
\label{alg:BNAO}
\begin{algorithmic}
\STATE \textbf{Input}: Initial candidate architectures set $\mathcal{X}$. Initial architectures set to be evaluated denoted as $\mathcal{X}_{eval}=\mathcal{X}$. Performances of architectures $\mathcal{S}=\emptyset$. Number of seed architectures $\mathcal{K}$. Number of training steps of the supernet $\mathcal{N}$. Step size $\eta$. Number of optimization iterations $\mathcal{L}$.
\FOR {$l= 1,\cdots, \mathcal{L}$}
\FOR {$i= 1,\cdots, \mathcal{N}$}
\STATE Sample an architecture $x\in \mathcal{X}_{eval}$ using Eqn.~(\ref{eqn:sample})
\STATE Train the supernet with the sampled architecture $x$
\ENDFOR
\STATE Evaluate all the candidate architectures $\mathcal{X}$ to obtain the dev set performances $\mathcal{S}_{eval}=\{s_x\}, \forall x \in \mathcal{X}_{eval}$. Enlarge $\mathcal{S}$: $\mathcal{S}=\mathcal{S}\bigcup \mathcal{S}_{eval}$.
\STATE Train the NAO model~(encoder, performance predictor and decoder) using $\mathcal{X}$ and $\mathcal{S}$.
\STATE Pick $\mathcal{K}$ architectures with top $\mathcal{K}$ performances among $\mathcal{X}$, forming the set of seed architectures $\mathcal{X}_{seed}$.
\STATE For $x \in \mathcal{X}_{seed}$, obtain a better representation $e_{x'}$ from $e_{x'}$ by applying gradient descent with step size $\eta$, based on trained encoder and performance predictor. Denote the set of enhanced representations as $E'=\{e_{x'}\}$.
\STATE Decode each $x'$ from $e_{x'}$ using decoder, set $\mathcal{X}_{eval}$ as the set of new architectures decoded out: $\mathcal{X}_{eval} = \{\mathcal{D}(e_{x'}), \forall e_{x'}\in E'\}$. Enlarge $\mathcal{X}$ as $\mathcal{X}= \mathcal{X}\bigcup \mathcal{X}_{eval}$.
\ENDFOR
\STATE \textbf{Output}: The architecture within $\mathcal{X}$ with the best performance
\end{algorithmic}
\end{algorithm}

\section{Experiments}
\label{sec:exp}

\subsection{Ranking Correlation Improvement}
\begin{table}[ht]
\centering
\begin{tabular}{cccc}
\hline
Training & Sampling & Pairwise \\
Epochs & Method & Accuracy(\%) \\
\hline
5 & Random sampling & 62 \\
5 & Proportional sampling & 69 \\
10 & Random sampling & 65 \\
10 & Proportional sampling & 72 \\
20 & Random sampling & 71 \\
20 & Proportional sampling & 76 \\
30 & Random sampling & 75 \\
30 & Proportional sampling & 77 \\
50 & Random sampling & 76 \\
50 & Proportional sampling & 78 \\
\hline
\end{tabular}
\caption{Pairwise accuracy of one-shot performance and ground-truth performance.}
\label{exp:ranking}
\end{table}
Firstly, to see whether our proposed proportional sampling improves the ranking correlation, we conduct experiments following the same setting as in Sec.~\ref{sec:imbalanced}. We train the $50$ architectures with and without proportional sampling, and compute the pairwise accuracy, as listed in Table~\ref{exp:ranking}. When equipped with proportional sampling, the accuracy increases by more than $10$ points.

\subsection{Stability Study}
\begin{table}[h]
\centering
\begin{tabular}{ccccccc}
\hline
Methods & ENAS & DARTS & NAO-WS & Balanced NAO \\
\hline
network 1 & 3.09 & 3.00 & 2.90 & \textbf{2.60}\\
\hline
network 2 & 3.00 & 2.82 & 2.96 & 2.63\\
\hline
network 3 & 2.97 & 3.02 & 2.98 & 2.65\\
\hline
network 4 & 2.79 & 2.96 & 3.02 & 2.71\\
\hline
network 5 & 3.07 & 3.00 & 3.05 & 2.77\\
\hline
mean      & 2.98 & 2.96 & 2.98 & \textbf{2.67}\\
\hline
\end{tabular}
\caption{Evaluation of stability. Scores are test error rate on CIFAR-10 of 5 architectures discovered. All models are trained with cutout. The first block lists the baseline one-shot NAS we compare to. The second block is our proposed Balanced NAO.}
\label{exp:stability}
\end{table}
Then, as we claim the stability to be a problem, we evaluate the stability of our proposed Balanced NAO and compare to baseline methods~(i.e., ENAS, DARTS, NAO-WS). Same to the experiments before, We evaluate $5$ architectures discovered by the algorithms and measure the mean performance. We search and test on CIFAR-10 and show the results in Table~\ref{exp:stability}. 

Compared to baseline methods, all the $5$ architectures perform better with an average test error rate of $2.67$\% and a best test error rate of \textbf{$2.60$\%}. Notably, we find that they perform well enough with lower variance, leading to better stability.

\subsection{Compared to Various NAS Methods}
We then compare our proposed Balanced NAO with other NAS methods, including NAS with or without the one-shot approach, to demonstrate the improvement of our approach. Table~\ref{tbl:cifar10} reports the test error rate, model size of several methods and our method. 

The first block lists several classical conventional NAS methods without the one-shot approach and trains each candidate architecture from scratch for several epochs. They achieved promising test error rate but require hundreds or thousands of GPU days to search, which is intractable for most researchers. The second block lists several main-stream one-shot NAS methods which largely reduce the cost to less than ten GPU days, but the test performance of most works are relatively inferior compared to conventional methods as we claimed before.

Note that our method in the last block achieves better performance than previous baseline one-shot NAS~(e.g., ENAS, DARTS, NAO-WS, SNAS), and is on par with conventional NAS~(e.g., NASNet and NAONet) under similar model size, while the latter is costly.

\subsection{Transfer to Large Scale Imagenet}
To further verify the transferability of our discovered architecture, we transfer it to large scale Imagenet dataset. All the training settings and details exactly follow DARTS~\cite{darts}. The results of our model and other models are listed in Table~\ref{tbl:imagenet} for comparison. 

Our model achieves a top-1 accuracy of $74.4$\%, which surpasses human designed models listed in the first block. We mainly compare our model to the models in the second block for fair comparison since they have roughly similar search space and apply the same transfer strategy from CIFAR to ImageNet. Our model surpasses many NAS methods in the second block and is on par with the best-performing work with minor gap. It is worth noting that NASNet-A, AmoebaNet-A, MnasNet and NAONet are all discovered with conventional NAS approaches that are not one-shot methods and the architectures are trained from scratch. Our model achieves better or almost the same performance compared to them. 

However, it is slightly worse than~\cite{singlepathnas}, ~\cite{singlepathoneshot},~\cite{proxylessnas} and~\cite{pcdarts} in the third blocks which directly search on ImageNet with more careful design and more complex search space.

\section{Conclusion}
In this paper, we find that due to the insufficient optimization problem in one-shot NAS, architectures of different sizes are imbalanced, which introduces incorrect rewards and leads to weak ranking correlation. This further misleads the search algorithm and leads to inferior performance against conventional NAS and weak stability. We therefore propose Balanced NAO by balancing the architectures via proportional sampling. Extensive experiments show that our proposed method shows significant improvements both on performance and stability. For future work, we would like to further investigate into one-shot NAS to improve its performance.

\bibliographystyle{ijcai20}
\bibliography{ijcai20}

\end{document}